\pdfoutput=1

\documentclass[11pt]{article}

\usepackage[]{emnlp2021}

\usepackage{times}
\usepackage{latexsym}

\usepackage[T1]{fontenc}

\usepackage[utf8]{inputenc}

\usepackage{microtype}

\usepackage{array}
\usepackage{booktabs}
\usepackage{amsmath}
\usepackage{graphicx}
\usepackage{hhline}
\usepackage{multirow}
\usepackage[symbol]{footmisc}
\usepackage{xspace}
\usepackage{stfloats}

\DeclareMathOperator*{\argmax}{arg\,max}

\newcommand{\method}{$\textrm{PMI}_\textrm{DC}$\xspace}
\newcommand{\prob}{\textsc{LM}\xspace}
\newcommand{\avgll}{\textsc{Avg}\xspace}
\newcommand{\unc}{\textsc{Unc}\xspace}
\newcommand{\calib}{\textsc{CC}\xspace}
\newcommand{\codeurl}{\url{https://github.com/peterwestuw/surface-form-competition}}

\newcommand\premise[1]{$[$\textcolor{blue}{#1}$]_\text{P}$}
\newcommand\uhypo[1]{$[$\textcolor{red}{#1}$]_\text{UH}$}
\newcommand\dpremise[1]{$[$#1$]_\text{DP}$}
\newcolumntype{P}[1]{>{\raggedright\arraybackslash}p{#1}}
\newcommand{\specialcellleft}[2][l]{%
\begin{tabular}[#1]{@{}l@{}}#2\end{tabular}
}

\title{Surface Form Competition: \\Why the Highest Probability Answer Isn't Always Right}

\author{\textsuperscript{=}Ari Holtzman\textsuperscript{1} \hspace{.1cm}
\textsuperscript{=}Peter West\textsuperscript{1,2} \\ \textbf{Vered Shwartz\textsuperscript{1,2}} \hspace{.1cm} \textbf{Yejin Choi\textsuperscript{1,2}} \hspace{.1cm} \textbf{Luke Zettlemoyer\textsuperscript{1}}  \\
  \textsuperscript{1}Paul G. Allen School of Computer Science \& Engineering, University of Washington\\
  \textsuperscript{2}Allen Institute for Artificial Intelligence\\
  \texttt{\{ahai,pawest\}@cs.washington.edu} \\
  }

\begin{document}

\maketitle


\begin{abstract}

\renewcommand\thefootnote{=}\footnotetext{Authors contributed equally}

\renewcommand*{\thefootnote}{\arabic{footnote}}
\setcounter{footnote}{0}

Large language models have shown promising results in \textit{zero-shot} settings \citep{brown2020language, radford2019language}. For example, they can perform multiple choice tasks simply by conditioning on a question and selecting the answer with the highest probability.

However, ranking by string probability can be problematic due to \textbf{surface form competition}—wherein different surface forms compete for probability mass, even if they represent the same underlying concept in a given context, e.g. ``computer'' and ``PC.'' Since probability mass is finite, this lowers the probability of the correct answer, due to competition from other strings that are valid answers (but not one of the multiple choice options).

\begin{figure}[t]
    \centering

    \includegraphics[width=\linewidth]{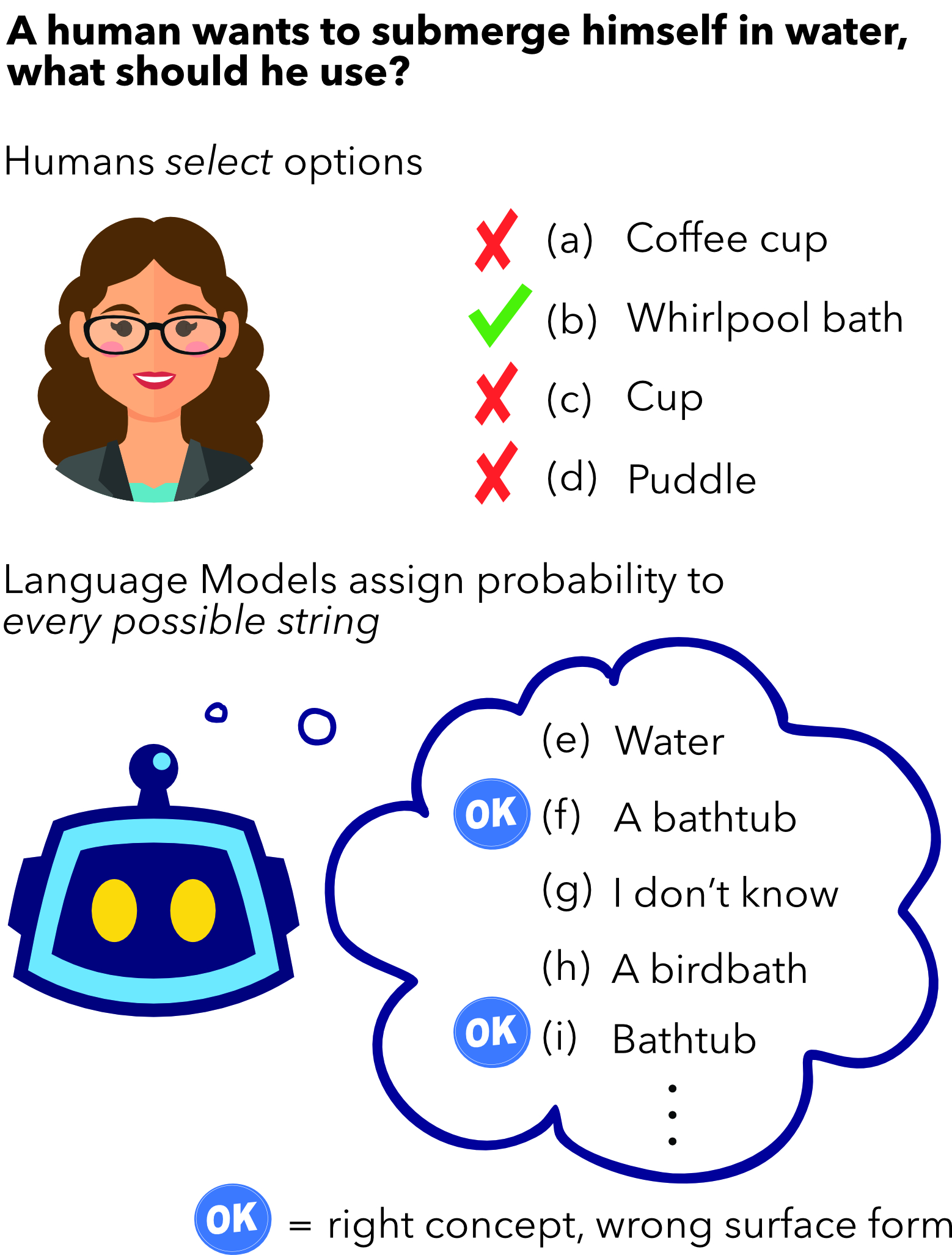}
    
    \caption{ While humans select from given options, language models implicitly assign probability to every possible string. This creates surface form competition between different strings that represent the same concept. Example from CommonsenseQA \cite{talmor2019commonsenseqa}.}
    \label{fig:1}
\end{figure} 

We introduce Domain Conditional Pointwise Mutual Information, an alternative scoring function that directly compensates for surface form competition by simply reweighing each option according to its a priori likelihood within the context of a specific task. It achieves consistent gains in zero-shot performance over both calibrated \cite{zhao2021calibrate} and uncalibrated scoring functions on all GPT-2 and GPT-3 models on a variety of multiple choice datasets.\footnote{Code is available at \codeurl}
\end{abstract}

\section{Introduction}

Despite the impressive results large pretrained language models have achieved in zero-shot settings \cite{brown2020language, radford2019language}, we argue that current work underestimates the zero-shot capabilities of these models on classification tasks. This is in large part due to \textbf{surface form competition}—a property of generative models that causes probability to be rationed between different valid strings, even ones that differ trivially, e.g., by capitalization alone.
Such competition can be largely removed by scoring choices according to Domain Conditional Pointwise Mutual Information \mbox{(\method)}, which reweighs scores by how much \textit{more} likely a hypothesis (answer) becomes given a premise (question) within the specific task domain.


Specifically, consider the example question (shown in Figure~\ref{fig:1}):  ``A human wants to submerge himself in water, what should he use?'' with multiple choice options ``Coffee cup'', ``Whirlpool bath'', ``Cup'', and ``Puddle.'' From the given options, ``Whirlpool bath'' is the only one that makes sense. Yet, other answers are valid and easier for a language model to generate, e.g., ``Bathtub'' and ``A bathtub.'' Since all surface forms compete for finite probability mass, allocating significant probability mass to ``Bathtub'' decreases the amount of probability mass assigned to ``Whirlpool bath.'' While the total probability of generating \textit{some correct answer} may be high (i.e., across all valid surface forms), only one of these is a listed option. This is particularly problematic here, because ``Whirlpool bath'' will be much lower probability than ``Bathtub,'' due to its rarity. More generally, methods that do not account for surface form competition will favor answers with fewer lexical paraphrases. 

\method factors out the probability of a specific surface form, by instead computing how much more probable a hypothesis is when conditioned on a premise. We use a \emph{domain premise} string to estimate the \emph{unconditional} probability of a hypothesis in a given domain. On CommonsenseQA, for example, we compute the probability of each answer option immediately following  the string ``? the answer is:'', and then divide the \textit{conditional} probability by this estimate to calculate \mbox{\method.} This scaling factor reweighs answer scores according to the surface form competition that is inherent to the domain or task, e.g. completions of the domain premise that are just inherently unlikely will be upweighted more. 
This allows us to directly measure how much an answer tells us about the question and vice versa (mutual information is symmetric, see \S\ref{sec:method}). 
Valid hypotheses no longer need to compete with each other: both ``Whirlpool bath'' and ``Bathtub '' will be considered reasonable answers to the question, and so both will attain a high score.

Extensive experiments show that \method consistently outperforms raw, normalized, and calibrated probability scoring methods on zero-shot multiple choice for more than a dozen datasets and it does so for every model in the GPT-2 and GPT-3 families (\S\ref{sec:exp}); this holds true across different possible prompts and in preliminary few-shot experiments as well. To better explain these gains, we use the distinct structure of the COPA dataset \cite{roemmele2011choice} to remove surface form competition entirely, showing that all methods perform well in this idealized setting~(\S\ref{exp:sfc}). Additionally, we analyze the only three datasets where \method does worse than other methods and put forward a hypothesis for why normalizing log probabilities works better than raw probabilities (\S\ref{sec:analysis}). We conclude with a discussion of how generative models should be used for selection tasks (\S\ref{sec:discussion}).

\section{Background and Related Work}
\label{sec:background}

\paragraph{Zero-shot vs. Few-Shot}
Zero-shot inference has long been of interest in NLP, Computer Vision, and ML in general \cite{socher2013recursive, guadarrama2013youtube2text, romera2015embarrassingly}. However, \citet{radford2019language} popularized the notion that language models have many zero-shot capabilities that can be discovered simply by prompting the model, e.g., placing ``TL;DR'' 
(internet slang for Too Long; Didn't Read)
at the end of a passage causes the model to generate a summary. Efficiently constructing the right prompt for a given task is difficult and has become an active area of research \cite{reynolds2021prompt, lu2021fantastically, shin-etal-2020-autoprompt, jiang2020can, jiang-etal-2020-know}.

\citet{brown2020language} demonstrated that few-shot learning without fine-tuning is possible with very large language models. Contemporary work has shown it is possible to get smaller models to exhibit few-shot learning behavior using fine-tuning \cite{hambardzumyan2021warp, gao2020making,  schick2020exploiting, schick2020few,schick2020s, shin-etal-2020-autoprompt}, an intermediate learning phase \cite{ye2021crossfit}, or calibration \cite{zhao2021calibrate}, though most assume access to a validation set \cite{perez2021true}.
Recent work suggests it may be possible to finetune language models in order to improve their zero-shot and few-shot capabilities on a large swathe of tasks \cite{wei2021finetuned, zhong2021adapting}.

\paragraph{Surface Form Competition}
When applying generative models to multiple choice problems, simply choosing the \emph{highest probability} answer becomes problematic due to 
different valid surface forms competing for probability.
Indeed, recent work in question answering has demonstrated the importance of considering all multiple choice options together \cite{khashabi2020unifiedqa}, rather than independently assigning each answer a score and simply choosing the highest. This is a difficult strategy to adapt to left-to-right generative language models, which implicitly choose between \emph{all} possible strings. Using unsupervised language models pretrained on relatively expansive corpora exacerbates surface form competition because such language models generate a much wider distribution than a given question answering dataset contains.



``What is the most populous nation in North America?'' Posed with this question, a language model such as GPT-3 can generate a correct response such as ``USA'', ''United States'', or ``United States of America'' with high probability. While correct strings like this all contribute to the probability of a correct generation, they may have vastly different probabilities: a common string ``United States'' will be much more likely than rarer forms like ``U.S. of A.''. In generative scenarios, as long as most of the probability mass goes to valid strings the generation is likely to be valid. This is not the case for multiple choice problems. Given two options, e.g., ``USA'' and ``Canada'', GPT-3 will choose the correct answer by probability. However, if we substitute out ``USA'' for ``U.S. of A.'', GPT-3 will assign higher probability to ``Canada'', a less likely answer conceptually, but a much more likely surface form.  Beyond this, incorrect generic answers such as ``I don't know'' are often assigned high probability, relegating the desired answers to the tail of the distribution where softmax is poorly calibrated \cite{holtzman2020curious}.



\paragraph{PMI} Work in dialogue has used PMI to promote diversity \cite{zhou2019unsupervised, yao2017towards, li2016diversity, mou2016sequence, tang2019target}. Recently, \citet{brown2020language} used a scoring function resembling \method for zero-shot question answering, though they only use the string ``A:'' as a prompt for the unconditional probability estimate, whereas we use a task-specific domain premise (see \S\ref{sec:method} for details). Furthermore, \citet{brown2020language} only report this scoring method on three datasets (ARC, OpenBookQA, and RACE, included here) out of the more than 20 tested and do not compare scores with their standard method, averaging log-likelihoods (\avgll in this work). In contrast, we report a comprehensive comparison on GPT-3 and GPT-2, as well as shedding light on the underlying issue of surface form competition in \S\ref{exp:sfc}.

\paragraph{Contextual Calibration} Recently, \citet{zhao2021calibrate} describe a new method for \textbf{calibrating} the probabilities of an LM using a learned affine transformation. Though geared towards few-shot learning, the authors devise a clever means of using ``content free inputs'' for zero-shot learning.  \citet{zhao2021calibrate} calibrate for three forms of bias: (1) majority label bias, (2) recency bias, and (3) common token bias. \method directly compensates for common token bias by dividing by the domain conditional probability of each answer, and performs superior to contextual calibration (\calib) in the majority of cases.

\begin{figure*}[t]
    \centering

    \includegraphics[width=\linewidth]{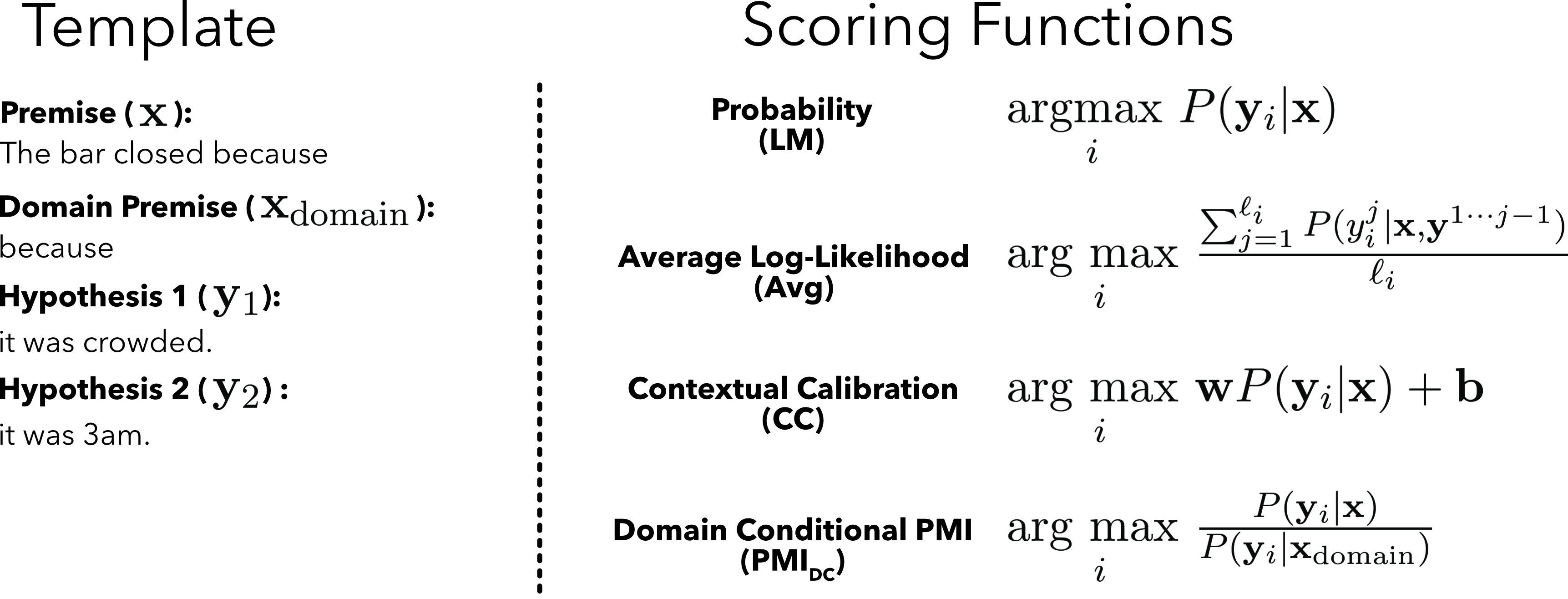}
    
\caption{An example from COPA \cite{roemmele2011choice} with the template we use as well as the scoring functions we test. \prob returns the highest probability option, while \avgll length-normalizes log-likelihoods and chooses the highest option. \method is a measurement of the mutual information between hypothesis and premise, intuitively how much $\mathbf{x}$ explains $\mathbf{y}_i$ and vice versa. \calib is an affine transform of \prob, where $\mathbf{w}$ and $\mathbf{b}$ are averaged over solutions that cause ``content free inputs'' to yield uniform scores over a given label set, see \citet{zhao2021calibrate}.}
    \label{fig:ex}
\end{figure*}

\paragraph{Prompt Sensitivity} Recent work highlights LM sensitivity to \emph{inputs}, and proposes to consider paraphrases of the prompt to overcome this \cite{davison-etal-2019-commonsense,jiang-etal-2020-know}, as well as noting that certain trigger tokens \cite{shin-etal-2020-autoprompt} can strongly effect the output of such models. In this work, we focus on the surface form of possible \emph{outputs}, but do also analyze robustness to different prompts in \S\ref{ssec:robustness}.

\paragraph{Interpreting Language Models} Language models tend to model selectional preferences and thematic fit \cite{pantel-etal-2007-isp,erk-etal-2010-flexible} rather than semantic plausibility \cite{wang-etal-2018-modeling}. Probability, possibility and plausibility are distinct \cite{van2006towards}, but reporting bias \cite{gordon2013reporting} means that language models only model what people are likely to write (on websites that are easily crawled). \method aims to adjust for these challenges to better measure the underlying agreement between language models and human judgements, but of course is still subject to the limits and biases of the language model used.
\section{Zero-shot Scoring Strategies}
\label{sec:method}

This paper does not define any new modeling or finetuning methods. Rather, we propose the broad use of \method scoring for any given model and prompt. \method compensates for the fact that different correct answers compete for probability, even though only one will be listed as the correct multiple choice option. 

We begin by describing the two most common methods currently in use.

\subsection{Standard Methods} Our first baseline is simply selecting the highest-probability option, e.g., baselines in \citet{zhao2021calibrate} and \citet{jiang-etal-2020-know}, which we refer to as \mbox{\prob.} Given a prompt $\mathbf{x}$ (e.g. ``The bar closed'') and a 
set of possible answers $\mathbf{y}_1, \cdots, \mathbf{y}_n$ (e.g. ``it was crowded.'', ``it was 3 AM.''), \prob is defined:
\begin{equation}
\argmax_i P(\mathbf{y}_i|\mathbf{x}).
\label{eqn:prob_def}
\end{equation}
However, using \emph{length normalized} log-likelihoods \cite{brown2020language} has become standard due to its superior performance, and is also commonly used in generation \cite{mao2019improving, oluwatobi-mueller-2020-dlgnet}. For causal language models, e.g., GPT-2 and GPT-3,  Equation~\ref{eqn:prob_def} can be decomposed:
\begin{equation*}
P(\mathbf{y}_i|\mathbf{x}) = \prod_{j=1}^{\ell_i} P(y_i^j|\mathbf{x},y_i^1,\cdots,y_i^{j-1})
\label{eqn:ll}
\end{equation*}
where $y_i^j$ is the $j$th token of $\mathbf{y}_i$ and $\ell_i$ is the number of tokens in $\mathbf{y}_i$. The \avgll strategy can thus be defined as:
\begin{equation*}
\argmax_i \frac{\sum^{\ell_i}_{j=1}\log P(y_i^j|\mathbf{x},\mathbf{y}^{1\cdots j-1})}{\ell_i}. \\
\label{eqn:avg_def}
\end{equation*}

\subsection{Domain Conditional PMI} Our core claim is that direct probability is not an adequate zero-shot scoring function due to surface form competition. A natural solution is to factor out the probability of specific surface forms, which is what Pointwise Mutual Information (PMI) does:
\begin{equation}
\textrm{PMI}(\mathbf{x},\mathbf{y}) = \log \frac{P(\mathbf{y}|\mathbf{x})}{P(\mathbf{y})} = \log \frac{P(\mathbf{x}|\mathbf{y})}{P(\mathbf{x})}.
\label{eqn:pmi_def}
\end{equation}
\noindent
In effect, this is how much more likely the hypothesis (``it was 3 AM.'') becomes given the premise (``The bar closed because''), see Figure~\ref{fig:ex} for the full example. 
In a multiple-choice setting—where the premise $\mathbf{x}$ does not change across hypotheses—this is proportional to $P(\mathbf{x}|\mathbf{y})$, i.e. the probability of the \emph{premise} given the \emph{hypothesis}. We call this scoring-by-premise and it is the reverse of \prob, $P(\mathbf{y}|\mathbf{x})$. We use scoring-by-premise to show the presence of surface form competition in \S\ref{exp:sfc}.

While Equation~\ref{eqn:pmi_def} estimates how related premise $\mathbf{x}$ is to hypothesis $\mathbf{y}$ in general, we found that estimates of $P(\mathbf{y})$ vary wildly. GPT-2 and GPT-3 are not trained to produce unconditional estimates of document excerpts, an issue which is exacerbated by the fact that many possible answers are extremely rare in a large scrape of public web pages. This causes the unconditional probability of such answers to be poorly calibrated for the purposes of a given task. 

We are specifically trying to measure $P(\mathbf{y})$ in a given domain, e.g., for the ``because'' relation in our running example, shown in Figures~\ref{fig:ex}~\&~\ref{fig:sfc}. To quantify this, we propose \textit{Domain~Conditional}~PMI:
\begin{align*}
\exp \textrm{PMI}_\textrm{DC}(\mathbf{x},\mathbf{y},\textrm{domain}) &= \frac{P(\mathbf{y}|\mathbf{x}, \textrm{domain})}{P(\mathbf{y}|\textrm{domain})} \\
&= \frac{P(\mathbf{y}|\mathbf{x}, \textrm{domain})}{P(\mathbf{y}|\mathbf{x}_\textrm{domain})}
\label{eqn:dcpmi_def}
\end{align*}
\noindent
or how much $\mathbf{x}$ tells us about $\mathbf{y}$ within a $\textrm{domain}$. 

Typically, $P(\mathbf{y}|\mathbf{x},\textrm{domain}) = P(\mathbf{y}|\mathbf{x})$ because the premise $\mathbf{x}$ typically implies the domain, e.g., ``The bar closed because'' sets the model up to predict an independent clause that is the cause of some event, without further representation of the domain.
In order to estimate $P(\mathbf{y}|\textrm{domain})$—the probability of seeing hypothesis $\mathbf{y}$ in a given $\textrm{domain}$—we use a short domain-relevant string $\mathbf{x}_\textrm{domain}$, which we call a ``domain premise'', usually just the ending of the conditional premise $\mathbf{x}$. For example, to predict a causal relation like in Figure~\ref{fig:ex} we use $\mathbf{x}_\textrm{domain}=\textrm{``because''}$ and thus divide by $P(\mathbf{y}|\textrm{because})$--how likely $y$ is to be a ``cause'' . For examples of each template see Appendix~\ref{sec:templates}.

\subsection{Non-standard Baselines}

\paragraph{Unconditional} We also compare to the unconditional (in-domain) estimate as a scoring function:
\begin{align*}
\argmax_i P(\mathbf{y}_i|\mathbf{x}_\textrm{domain}).
\end{align*}
We refer to this as \unc. It ignores the premise completely, only using a domain premise $\mathbf{x}_\textrm{domain}$ (e.g., using $P(\mathbf{y}|\textrm{because})$ as the score). Yet, it is sometimes competitive, for instance on BoolQ \cite{clark2019boolq}. \unc is a sanity check on whether zero-shot inference is actually using the information in the question to good effect.

\paragraph{Contextual Calibration} Finally, we compare to the reported zero-shot numbers of \citet{zhao2021calibrate}.  \textit{Contextual Calibration} adjusts \prob with an affine transform to make a closed set of answers equally likely in the absence of evidence. Contextual Calibration thus requires computing matrices $\mathbf{w}$ and $\mathbf{b}$ for a number of ``content free inputs'' and then averaging these weights, see \citet{zhao2021calibrate} for details. In contrast, \method requires nothing but a human-written template (as all zero-shot methods do, including Contextual Calibration), can be computed as the difference of two log probabilities, and is naturally applicable to  datasets where the set of valid answers varies between questions.

\section{Multiple Choice Experiments}
\label{sec:exp}



\subsection{Setup}
\label{exp-setup}
We use GPT-2 via the HuggingFace Transformers library \cite{wolf-etal-2020-transformers} and GPT-3 via OpenAI's beta API.\footnote{\url{https://beta.openai.com/}} We do not finetune any models, nor do we alter their output. 
See Appendix~\ref{sec:templates} for examples from each dataset in our templated format.

\subsection{Datasets}
We report results on 16 splits of 13 datasets, and briefly describe each dataset here.

\paragraph{Continuation} These datasets require the model to select a continuation to previous text, making them a natural way to test language models. Choice of Plausible Alternatives (COPA) \cite{roemmele2011choice} asks for cause and effect relationships, as shown in Figure~\ref{fig:ex}. StoryCloze (SC) \cite{mostafazadeh2017lsdsem} gives the model a choice between two alternative endings to 5 sentence stories. Finally, HellaSwag (HS) \cite{zellers2019hellaswag} uses GPT-2 to generate, BERT to filter, and crowd workers to verify possible continuations to a passage. Following previous work \cite{brown2020language} we report development set numbers for COPA and HS.

\paragraph{Question Answering} RACE-M \& -H (R-M \& R-H) \cite{lai2017race}  are both drawn from English exams given in China, the former being given to Middle Schoolers and the latter to High Schoolers. Similarly, ARC Easy \& Challenge (ARC-E \& ARC-C) \cite{clark2018think}  are standardized tests described as ``natural, grade-school science questions,'' with the ``Easy'' split found to be solvable by either a retrieval or word co-occurrence system, and the rest of the questions put in the ``Challenge'' split. Open Book Question Answering (OBQA) \cite{mihaylov2018can} is similar to both of these, but was derived using (and intended to be tested with) a knowledge source (or ``book'') available; we do not make use of the given knowledge source, following \citet{brown2020language}. Finally, CommonsenseQA (CQA) \cite{talmor2019commonsenseqa} leverages \textsc{ConceptNet}\hspace{2pt}\cite{speer2017conceptnet} to encourage crowd workers to write questions with challenging distractors. We report development set numbers on CQA because their test set is not public.

\begin{table*}[t]
\small
\setlength{\tabcolsep}{2.5pt}
    \centering
    \caption*{Multiple Choice Accuracy on GPT-3}
    \begin{tabular}{l|ccccc|cccc|cccc|ccccc}
     Params. & \multicolumn{5}{c}{2.7B} & \multicolumn{4}{c}{6.7B} & \multicolumn{4}{c}{13B} & \multicolumn{5}{c}{175B} \\
    &  Unc  &  LM  &  Avg  &  $\textrm{PMI}_\textrm{DC}$  &  CC  &  Unc  &  LM  &  Avg  &  $\textrm{PMI}_\textrm{DC}$  &  Unc  &  LM  &  Avg  &  $\textrm{PMI}_\textrm{DC}$  &  Unc  &  LM  &  Avg  &  $\textrm{PMI}_\textrm{DC}$  &  CC  \\
    \midrule
    COPA  &  54.8  &  68.4  &  68.4  &  \textbf{74.4}  &  -  &  56.4  &  75.8  &  73.6  &  \textbf{77.0}  &  56.6  &  79.2  &  77.8  &  \textbf{84.2}  &  56.0  &  85.2  &  82.8  &  \textbf{89.2}  &  -  \\
    SC  &  50.9  &  66.0  &  68.3  &  \textbf{73.1}  &  -  &  51.4  &  70.2  &  73.3  &  \textbf{76.8}  &  52.0  &  74.1  &  77.8  &  \textbf{79.9}  &  51.9  &  79.3  &  83.1  &  \textbf{84.0}  &  -  \\
    HS  &  31.1  &  34.5  &  \textbf{41.4}  &  34.2  &  -  &  34.7  &  40.8  &  \textbf{53.5}  &  40.0  &  38.8  &  48.8  &  \textbf{66.2}  &  45.8  &  43.5 &  57.6  &  \textbf{77.2}  &  53.5  &  -  \\
    R-M  &  22.4  &  37.8  &  42.4  &  \textbf{42.6}  &  -  &  21.2  &  43.3  &  45.9  &  \textbf{48.5}  &  22.9  &  49.6  &  50.6  &  \textbf{51.3}  &  22.5  &  55.7  &  \textbf{56.4}  &  55.7  &  -  \\
    R-H  &  21.4  &  30.3  &  32.7  &  \textbf{36.0}  &  -  &  22.0  &  34.8  &  36.8  &  \textbf{39.8}  &  22.9  &  38.2  &  39.2  &  \textbf{42.1}  &  22.2  &  42.4  &  43.3  &  \textbf{43.7}  &  -  \\
    ARC-E &  31.6  &  \textbf{50.4}  &  44.7  &  44.7  &  -  &  33.5  &  \textbf{58.2}  &  52.3  &  51.5  &  33.8  &  \textbf{66.2}  &  59.7  &  57.7  &  36.2  &  \textbf{73.5}  &  67.0  &  63.3  &  -  \\
    ARC-C &  21.1  &  21.6  &  25.5  &  \textbf{30.5}  &  -  &  21.8  &  26.8  &  29.8  &  \textbf{33.0}  &  22.3  &  32.1  &  34.3  &  \textbf{38.5}  &  22.6  &  40.2  &  43.2  &  \textbf{45.5}  &  -  \\
    OBQA  &  10.0  &  17.2  &  27.2  &  \textbf{42.8}  &  -  &  11.4  &  22.4  &  35.4  &  \textbf{48.0}  &  10.4  &  28.2  &  41.2  &  \textbf{50.4}  &  10.6  &  33.2  &  43.8  &  \textbf{58.0}  &  - \\
    CQA  &  15.9  &  33.2  &  36.0  &  \textbf{44.7}  &  -  &  17.4  &  40.0  &  42.9  &  \textbf{50.3}  &  16.4  &  48.8  &  47.9  &  \textbf{58.5}  &  16.3  &  61.0  &  57.4  &  \textbf{66.7}  &  - \\
    \midrule
    BQ  &  \textbf{62.2}  &  58.5  &  58.5  &  53.5  &  -  &  37.8  &  \textbf{61.0}  &  \textbf{61.0}  &  \textbf{61.0}  &  \textbf{62.2}  &  61.1  &  61.1  &  60.3  &  37.8  &  62.5  &  62.5  &  \textbf{64.0}  &  - \\
    RTE  &  47.3  &  48.7  &  48.7  &  \textbf{51.6}  &  49.5  &  52.7  &  \textbf{55.2}  &  \textbf{55.2}  &  48.7  &  52.7  &  52.7  &  52.7  &  \textbf{54.9}  &  47.3  &  56.0  &  56.0  &  \textbf{64.3}  &  57.8  \\
    CB  &  08.9  &  51.8  &  51.8  &  \textbf{57.1}  &  50.0  &  08.9  &  33.9  &  33.9  &  \textbf{39.3}  &  08.9  &  \textbf{51.8}  &  \textbf{51.8}  &  50.0  &  08.9  &  48.2  &  48.2  &  \textbf{50.0}  &  48.2 \\
    SST-2  &  49.9 & 53.7 & 53.76 & \textbf{72.3}  &  71.4  &  49.9 & 54.5 & 54.5 & \textbf{80.0}  &  49.9 & 69.0 & 69.0 & \textbf{81.0}  &  49.9 & 63.6 & 63.6 & 71.4  &  \textbf{75.8} \\
SST-5  &  18.1 & 20.0 & 20.4 & \textbf{23.5} &  -  &  18.1 & 27.8 & 22.7 & \textbf{32.0} &  18.1 & 18.6 & \textbf{29.6} & \textbf{19.1} & 17.6 & 27.0 & 27.3 & \textbf{29.6} &  - \\
    AGN  &  25.0  &  69.0  &  69.0  &  \textbf{67.9}  &  63.2  &  25.0  &  \textbf{64.2}  &  \textbf{64.2}  &  57.4  &  25.0  &  69.8  &  69.8  &  \textbf{70.3}  &  25.0  &  \textbf{75.4}  &  \textbf{75.4}  &  74.7  &  73.9 \\
    TREC  &  13.0  &  29.4  &  19.2  &  \textbf{57.2}  &  38.8  &  22.6  &  30.2  &  22.8  &  \textbf{61.6}  &  22.6  &  \textbf{34.0}  &  21.4  &  32.4  &  22.6  &  47.2  &  25.4  &  \textbf{58.4}  &  57.4 \\
    \end{tabular}
    \caption{Comparison of scoring algorithms when using GPT-3 for zero-shot inference on multiple choice questions.}
    \label{tab:mc-3}
\end{table*}
\begin{table}[t]
\setlength{\tabcolsep}{4pt}
\caption*{Percent of Ties or Wins by Method}
    \centering
    \begin{tabular}{l|rrrrr}
    
    Method  &  Unc  &  LM  &  Avg  &  $\textrm{PMI}_\textrm{DC}$  &  CC  \\
    \midrule 
    125M  &  12.50  &  6.25  &  12.50  &  \textbf{68.75}  &  -  \\
    350M  &  6.25  &  18.75  &  12.50  &  \textbf{68.75}  &  -   \\
    760M  &  6.25  &  6.25  &  12.50  &  \textbf{75.00}  &  -  \\
    1.6B  &  6.25  &  12.50  &  12.50  &  \textbf{80.00}  &  20.00  \\
    \midrule 
    2.7B  &  6.25  &  6.25  &  6.25  &  \textbf{86.66}  &  0.00  \\
    6.7B  &  6.25  &  25.00  &  25.00  &  \textbf{75.00}  &  -  \\
    13B  &  6.25  &  18.75  &  18.75  &  \textbf{68.75}  &  -  \\
    175B  &  6.25  &  12.50  &  18.75  &  \textbf{62.50}  &  6.25  \\
    \end{tabular}
    \caption{Percentage of datasets that a given method produced the best score or was tied with other methods, aggregated over each model size. The first four rows use GPT-2 (full data available in the Appendix), while the final four rows use GPT-3 and summarize data from Table~\ref{tab:mc-3}. Since ties are included, rows sometimes sum to more than 100. CC is only measured on the 5 datasets we use where \citet{zhao2021calibrate} also report accuracies.}
    \label{tab:summ}
\end{table}

\paragraph{Open Set vs. Closed Set Datasets} The above datasets are all ``open set'' in that multiple choice answers may be any string. Below we describe ``closed set'' datasets with a fixed set of answers. 

\paragraph{Boolean Question Answering} 
BoolQ (BQ) \cite{clark2019boolq}  poses yes/no (i.e. Boolean) questions based on a multi-sentence passage.

\paragraph{Entailment} Entailment datasets focus on the question of whether a hypothesis sentence B is entailed by a premise sentence A. Recognizing Textual Entailment (RTE) \cite{dagan2005pascal} requires predicting an ``entailment'' or ``contradiction'' label while Commitment Bank (CB) \cite{de2019commitmentbank} adds a ``neutral'' label. Following previous work \cite{brown2020language} we report development set numbers for both RTE and CB.

\paragraph{Text Classification} We consider three more complex classification datasets: SST-2 \& -5  \cite{socher2013recursive} for various granularities of sentiment classification, AG's News \cite{zhang2015character} (AGN) for topic classification, and TREC \cite{li2002learning} for question classification.

\subsection{Results}
We report zero-shot results for GPT-3 in Table~\ref{tab:mc-3}, with GPT-2 results available in Appendix~\ref{sec:gpt2}. A summarized view is shown in Table~\ref{tab:summ}, which aggregates the percentage of splits where a given method achieves the best score or ties for first-place. In this summarized view it is clear that \method consistently outperforms other scoring methods when assessed over a variety of datasets. The smallest margin (in number of datasets won or tied) between \method and the best competing method is on GPT-3 175B with \avgll, but that margin is over 40 percentage points. This does not imply that \method is \textit{always} better or that it will be better by a large margin, though it often is. It does  suggest that \method is a significantly better bet on a new dataset.

\begin{table}[t]
\caption*{Prompt Robustness on SST-2}
    \centering
    \begin{tabular}{l|lll}
    Method  &  Unc  & LM  & $\textrm{PMI}_\textrm{DC}$ \\
    \midrule 
    125M  & $\textrm{49.9}_\textrm{~0}$ & $\textrm{56.8}_\textrm{~7.3}$  & $\textrm{\textbf{58.8}}_\textrm{~7.6}$  \\
    350M  & $\textrm{49.9}_\textrm{~0}$ & $\textrm{58.0}_\textrm{~11.3}$ & $\textrm{\textbf{60.3}}_\textrm{~11.4}$ \\
    760M  & $\textrm{49.9}_\textrm{~0}$ & $\textrm{57.0}_\textrm{~9.2}$  & $\textrm{\textbf{67.7}}_\textrm{~13.4}$ \\
    1.6B  & $\textrm{49.9}_\textrm{~0}$ & $\textrm{57.3}_\textrm{~8.2}$  & $\textrm{\textbf{69.8}}_\textrm{~13.3}$ \\
    \midrule 
    2.7B  & $\textrm{49.9}_\textrm{~0}$ & $\textrm{56.1}_\textrm{~9.0}$  & $\textrm{\textbf{66.2}}_\textrm{~15.7}$ \\
    6.7B  & $\textrm{49.9}_\textrm{~0}$ & $\textrm{59.5}_\textrm{~10.7}$ & $\textrm{\textbf{67.9}}_\textrm{~13.6}$ \\
    13B   & $\textrm{49.9}_\textrm{~0}$ & $\textrm{63.0}_\textrm{~14.9}$ & $\textrm{\textbf{71.7}}_\textrm{~16.1}$ \\
    175B  & $\textrm{49.9}_\textrm{~0}$ & $\textrm{72.5}_\textrm{~15.7}$ & $\textrm{\textbf{74.8}}_\textrm{~14.0}$ \\
    \end{tabular}
    \caption{The mean and standard deviations over the 15 templates considered for SST-2 in \cite{zhao2021calibrate}. \avgll is excluded, as it is equivalent to \prob since all the given templates use single-token answers.}
    \label{tab:robustness}
\end{table}
\begin{table}[t]
\small
\setlength{\tabcolsep}{1.5pt}
\caption*{4-shot Inference Results}
 \centering
 \begin{tabular}{l|lll|llll}
  \multicolumn{4}{c}{\hspace{15pt}SST-2} & \multicolumn{4}{c}{CQA} \\
 Method & Unc & LM & $\textrm{PMI}_\textrm{DC}$ & Unc & LM & Avg & $\textrm{PMI}_\textrm{DC}$ \\
 \midrule 
 125M & $\textrm{49.9}_\textrm{~0}$  & $\textrm{63.6}_\textrm{~7.4}$  & $\textrm{\textbf{71.7}}_\textrm{~5.1}$ & $\textrm{15.5}_\textrm{~0}$ & $\textrm{29.9}_\textrm{~1.6}$ & $\textrm{32.7}_\textrm{~1.4}$ & $\textrm{\textbf{38.3}}_\textrm{~1.7}$ \\
 350M  & $\textrm{49.9}_\textrm{~0}$  & $\textrm{76.3}_\textrm{~13.8}$  & $\textrm{\textbf{76.4}}_\textrm{~8.1}$ & $\textrm{16.5}_\textrm{~0}$ & $\textrm{37.6}_\textrm{~2.3}$ & $\textrm{40.4}_\textrm{~2.3}$ & $\textrm{\textbf{45.7}}_\textrm{~2.4}$ \\
 760M  & $\textrm{49.9}_\textrm{~0}$  & $\textrm{85.9}_\textrm{~7.2}$  & $\textrm{\textbf{87.1}}_\textrm{~3.0}$ & $\textrm{16.1}_\textrm{~0}$ & $\textrm{41.5}_\textrm{~2.6}$ & $\textrm{42.4}_\textrm{~2.5}$ & $\textrm{\textbf{47.0}}_\textrm{~1.5}$ \\
 1.6B  & $\textrm{49.9}_\textrm{~0}$  & $\textrm{85.4}_\textrm{~1.7}$  & $\textrm{\textbf{89.4}}_\textrm{~4.0}$ &  $\textrm{16.0}_\textrm{~0}$ & $\textrm{46.2}_\textrm{~1.5}$ & $\textrm{47.7}_\textrm{~1.9}$ & $\textrm{\textbf{52.3}}_\textrm{~2.1}$ \\
 \midrule
 2.7B  & $\textrm{49.9}_\textrm{~0}$  & $\textrm{\textbf{88.1}}_\textrm{~4.9}$  & $\textrm{87.7}_\textrm{~5.5}$ & $\textrm{16.6}_\textrm{~0}$ & $\textrm{43.0}_\textrm{~1.7}$ & $\textrm{45.6}_\textrm{~1.9}$ & $\textrm{\textbf{50.4}}_\textrm{~1.1}$ \\
 6.7B  & $\textrm{49.9}_\textrm{~0}$  & $\textrm{\textbf{92.9}}_\textrm{~2.1}$  & $\textrm{79.8}_\textrm{~6.9}$ & $\textrm{16.9}_\textrm{~0}$ & $\textrm{52.3}_\textrm{~1.4}$ & $\textrm{53.4}_\textrm{~1.0}$ & $\textrm{\textbf{56.5}}_\textrm{~1.6}$ \\
 13B  & $\textrm{49.9}_\textrm{~0}$  & $\textrm{85.4}_\textrm{~9.0}$  & $\textrm{\textbf{86.9}}_\textrm{~7.5}$ & $\textrm{16.7}_\textrm{~0}$ & $\textrm{58.4}_\textrm{~2.0}$ & $\textrm{59.3}_\textrm{~1.5}$ & $\textrm{\textbf{63.4}}_\textrm{~1.4}$ \\
 175B  & $\textrm{49.9}_\textrm{~0}$  & $\textrm{89.9}_\textrm{~5.5}$  & $\textrm{\textbf{95.5}}_\textrm{~0.7}$ & $\textrm{16.5}_\textrm{~0}$ & $\textrm{69.1}_\textrm{~1.9}$ & $\textrm{69.4}_\textrm{~0.8}$ & $\textrm{\textbf{72.0}}_\textrm{~0.9}$ \\
 \end{tabular}
 \caption{The mean and standard deviation for 5 randomly sampled sets of 4 examples used for few-shot inference. We include a closed answer dataset (SST-2) and an open answer dataset (CQA). For SST-2 \avgll is equivalent to \prob due to using single-token answers.}
 \label{tab:fewshot}
\end{table}

\subsection{Robustness}
\label{ssec:robustness}

To verify that these trends hold across different prompts, we report the mean and standard deviation over the fifteen different prompts considered in \cite{zhao2021calibrate} for SST-2. Table~\ref{tab:robustness} shows, \method always maintains the highest mean, often by a hefty margin. Scores are lower than in Table~\ref{tab:mc-3} because many of the prompts used are optimized for few-shot rather than zero-shot scoring.

\subsection{Few-shot}
\label{ssec:fewshot}

While our focus in this paper is on zero-shot scoring, \method is just as applicable to few-shot scenarios. In Table~\ref{tab:fewshot} we report 4-shot results on one closed set dataset (SST-2) and one open set dataset (CQA). We show the mean of 5 randomly sampled sets of 4 examples that are used to prime the model for the task, along with standard deviations. The overall trend on both datasets clearly favors \method, though \prob is superior for two models on SST-2. 

\section{Removing Surface Form Competition}
\label{exp:sfc}


What if we used the probability of the \textit{premise} given the \textit{hypothesis},  $P(\mathbf{x}|\mathbf{y}_i)$, instead? While we are still measuring the probability of a surface form (e.g. ``the bar closed.''), it is the \textit{same} surface form across different options (``It was crowded so'', ``It was 3 AM so''), eliminating the surface form competition. $\mathbf{y}_i$ and $\mathbf{y}_i'$ can now both attain high scores if they are both correct answers, by causing $\mathbf{x}$ to be likely. We call this scoring-by-premise.

\begin{figure*}[t]
    \centering

    \includegraphics[width=\linewidth]{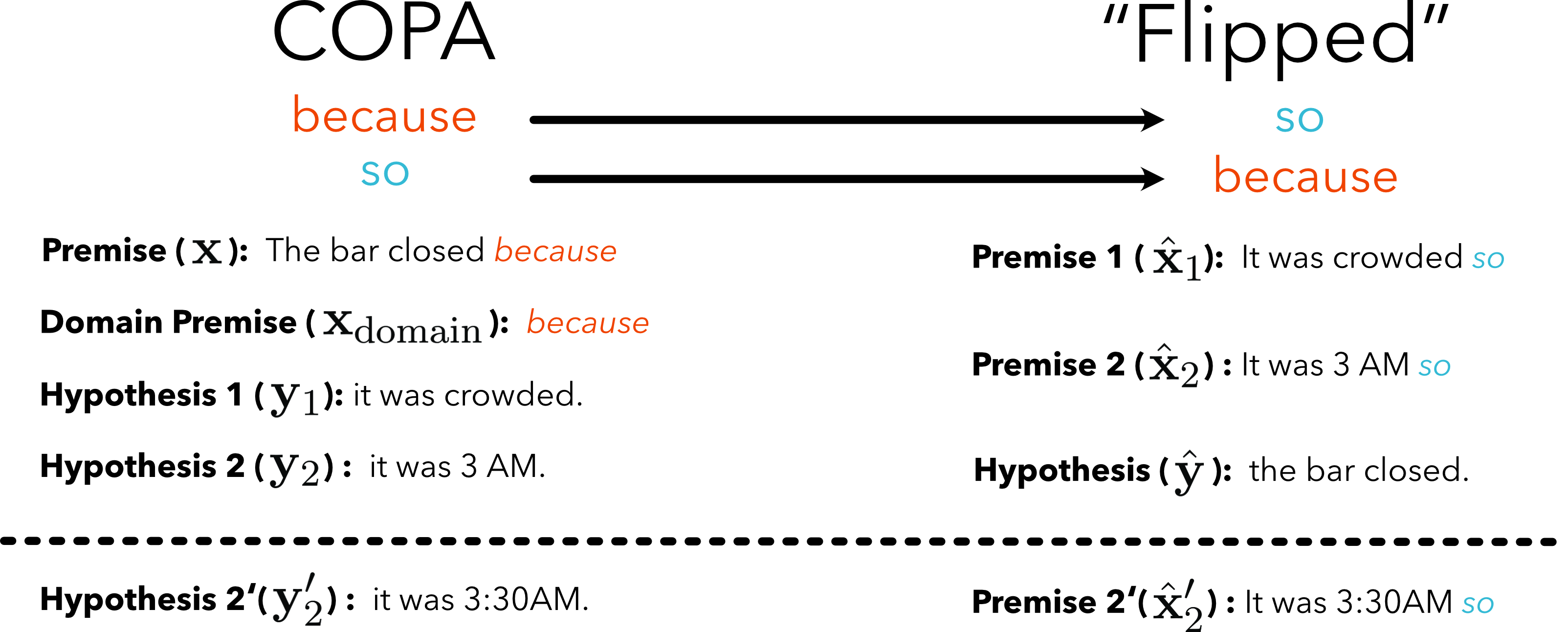}
    
    \caption{In \S\ref{exp:sfc} we experiment with
    with flipping the premise and hypothesis so that the highest probability \textit{premise} is chosen as the answer, i.e. scoring-by-premise. The transformation above the dashed line shows the experimental setup used in \S\ref{ssec:flipped-results}, while the extra distractor below the dashed line is used for illustrative purposes in \S\ref{ssec:flipped-example}.
    }
    \label{fig:sfc}
\end{figure*}

Causal language models like GPT-3 cannot measure this directly, because they are only capable of conditioning on past tokens to predict future tokens.
We exploit the structure of the COPA dataset to create ``COPA Flipped'' via a simple transformation, shown in Figure~\ref{fig:sfc}. COPA consists of cause and effect pairs (CAUSE \textit{so} EFFECT, and EFFECT \textit{because} CAUSE). In the original dataset, whatever comes second (either CAUSE or EFFECT) has two options that a model must choose between. These can be reversed by switching CAUSE and EFFECT, then substituting the natural inverse relation (``because''$\xrightarrow{}$``so'' and ``so''$\xrightarrow{}$``because'' ).

\subsection{Results}
\label{ssec:flipped-results}

Table~\ref{tab:sfc} shows scores on COPA and COPA Flipped side-by-side. On COPA Flipped everything except \unc produces the \emph{exact} same result. This is because flipping the hypothesis and premise means that it's the \textit{context} that changes and not the \textit{continuation}. \prob, \avgll, and \method only differ from each other over different continuations, not over different contexts for the same continuation.

\begin{table}[t]
\caption*{Removing Surface Form Competition}
\small
\setlength{\tabcolsep}{2.8pt}
    \centering
    \begin{tabular}{l|cccc|cccc}
\multicolumn{5}{c}{\hspace{40pt} COPA} & \multicolumn{4}{c}{ COPA Flipped} \\
    Method  &  Unc  &  LM  &  Avg  &  $\textrm{PMI}_\textrm{DC}$  &  Unc  &  LM  &  Avg  &  $\textrm{PMI}_\textrm{DC}$  \\
    \midrule 
    125M  &  56.4  &  61.0  &  63.2  &  62.8  &  50.0 &  63.2  &  63.2  &  63.2  \\
    350M  &  55.8  &  67.0  &  66.0  &  70.0  &  50.0 &  66.4  &  66.4  &  66.4  \\
    760M  &  55.6  &  69.8  &  67.6  &  69.4  &  50.0 &  70.8  &  70.8  &  70.8  \\
    1.6B  &  56.0  &  69.0  &  68.4  &  71.6  &  50.0 &  73.0  &  73.0  &  73.0  \\
    \midrule
    2.7B  &  54.8  &  68.4  &  68.4  &  74.4  &  50.0 &  68.4  &  68.4  &  68.4  \\
    6.7B  &  56.4  &  75.8  &  73.6  &  77.0  &  50.0  &  76.8  &  76.8  &  76.8  \\
    13B  &  56.6  &  79.2  &  77.8  &  84.2  &  50.0 &  79.0  &  79.0  &  79.0  \\
    175B  &  56.0  &  85.2  &  82.8  &  89.2  &  50.0 &  83.6  &  83.6  &  83.6
    \end{tabular}
    \caption{\prob does better on COPA Flipped than COPA because surface form competition is removed when scoring-by-premise, see \S\ref{exp:sfc}. Methods that don't directly adjust for competing surface forms (\prob and \mbox{\avgll\hspace{-3pt})} have the same score as \method on COPA Flipped.}
    \label{tab:sfc}
\end{table}

On COPA Flipped all methods generally perform similarly to \method on the unflipped version. This is because surface form competition has been eradicated: we are measuring how well different prefixes condition a model to predict a fixed continuation rather than which continuation is highest probability. Unlike \prob, where different answers compete for probability, in COPA Flipped it only matters how likely each answer can make the question. This is not subject to surface form competition because there is only one string being so scored, so it is not competing with any other strings for probability mass.

Not all datasets are so easily flippable, so manually flipping individual questions to remove surface form competition is not a generally applicable strategy. Luckily, \method is symmetric:
\begin{align*}
&\argmax_i \frac{P(\mathbf{y}_i|\mathbf{x},\textrm{domain})}{P(\mathbf{y}_i|\textrm{domain})} \\= &\argmax_i \frac{P(\mathbf{x}|\mathbf{y}_i,\textrm{domain})}{P(\mathbf{x}|\textrm{domain})} \\
= &\argmax_i P(\mathbf{x}|\mathbf{y}_i,\textrm{domain})
\end{align*}
\noindent
In theory,  the answer selected by \method should be the same between COPA and COPA Flipped as PMI is symmetric, though we expect some differences due to ``so'' and ``because'' not being perfect inverses and shuffled references. Thus, \method does better on COPA than COPA Flipped, likely due to more natural phrasing in the original dataset.

These results suggest that surface form competition is the primary cause of the depressed performance of \prob and \avgll in comparison to \method.

\subsection{In-depth Example}
\label{ssec:flipped-example}

\paragraph{Scoring-by-Premise Improves \prob} Figure~\ref{fig:sfc} shows an example of transforming one question from COPA to COPA Flipped. In the example depicted, when we use GPT-3 to calculate $P$, we get:
\begin{align*}
    P(\mathbf{y}_1|\mathbf{x}) > P(\mathbf{y}_2|\mathbf{x})
\end{align*}
which is wrong, since bars usually close at fixed, late-night closing times, rather than because of being overcrowded. However we also find that 
\begin{align*}
     P(\hat{\mathbf{y}}|\hat{\mathbf{x}}_2) &> P(\hat{\mathbf{y}}|\hat{\mathbf{x}}_1) \\
    \frac{P(\mathbf{y}_2|\mathbf{x})}{P(\mathbf{y}_2|\mathbf{x}_\textrm{domain})} &> \frac{P(\mathbf{y}_1|\mathbf{x})}{P(\mathbf{y}_1|\mathbf{x}_\textrm{domain})}
\end{align*}
indicating that scoring-by-premise causes the right answer to be selected and that \method successfully simulates scoring by premise in this example.

\paragraph{Stability Over Valid Answers} To see how scoring-by-premise allows multiple correct options to achieve high scores, consider the slightly perturbed $\mathbf{y}'_2$ and $\hat{\mathbf{x}}_2'$ in Figure~\ref{fig:sfc}. The inequalities shown above still hold when substituting \hspace{50pt} $\mathbf{y}_2\rightarrow\mathbf{y}_2'$ and $\hat{\mathbf{x}}_2\rightarrow\hat{\mathbf{x}}_2'$:
\begin{align*}
    P(\mathbf{y}_1|\mathbf{x}) &> P(\mathbf{y}_2'|\mathbf{x}) \\
     P(\hat{\mathbf{y}}|\hat{\mathbf{x}}_2') &> P(\hat{\mathbf{y}}|\hat{\mathbf{x}}_1) \\
    \frac{P(\mathbf{y}_2'|\mathbf{x})}{P(\mathbf{y}_2'|\mathbf{x}_\textrm{domain})} &> \frac{P(\mathbf{y}_1|\mathbf{x})}{P(\mathbf{y}_1|\mathbf{x}_\textrm{domain})}
\end{align*}
with the key difference that the conditional probability of $\mathbf{y}'_2$ is much lower:
\begin{align*}
    \log{P(\mathbf{y}_2|\mathbf{x})} &\approx -16 \\ \log{P(\mathbf{y}_2'|\mathbf{x})} &\approx -20 
\end{align*}
This is undesirable, as both $\mathbf{y}_2$ and $\mathbf{y}'_2$ are correct answers with similar meanings. Yet, when scoring-by-premise  the conditional probability of $\hat{\mathbf{y}}$ is stable when substituting $\hat{\mathbf{x}}_2\rightarrow\hat{\mathbf{x}}_2'$:
\begin{align*}
    \log{P(\hat{\mathbf{y}}|\hat{\mathbf{x}}_2)} &\approx -12 \\ 
    \log{P(\hat{\mathbf{y}}|\hat{\mathbf{x}}_2')} &\approx -12 
\end{align*}

This suggests that eliminating surface form competition allows different correct answers to score well, as they are no longer competing for probability mass. Specifically, ``it was 3 AM'' and ``it was 3:30AM'' score wildly differently in COPA but nearly identically in COPA Flipped.

\section{Analysis}
\label{sec:analysis} 

\paragraph{Failure Cases} 
 There are three datasets where \method does not consistently outperform other methods: HellaSwag, ARC Easy, and BoolQ. Surprisingly, each is dominated by a different method.

HellaSwag is most amenable to \avgll. On examination we find that HellaSwag is more focused on the \textit{internal coherence} of the hypotheses, 
 rather than \textit{external coherence}, i.e. how much a premise and hypothesis match. This is likely due to HellaSwag being generated by GPT-2 \cite{radford2019language} and filtered with BERT, as it contains relatively on-topic but intrinsically strange hypotheses that humans can distinguish from natural data.

ARC Easy yields the highest scores to \prob, i.e., selecting the highest probability option.
\citet{clark2018think} note that ARC Easy questions can be solved by a retrieval or word co-occurrence baseline, while examples that were answered incorrectly by both were put into the Challenge split. This suggests a bias towards a priori likely phrases. Manual inspection reveals many stock answers, e.g., ``[clouds are generated when] ocean water evaporates and then condenses in the air,'' supporting our hypothesis. 

Finally, BoolQ, a reading comprehension dataset in which all answers are either ``yes'' or ``no'', is best solved by an unconditional baseline. This is because the dataset presents truly complex questions that require more reasoning than GPT-2 or 3 are capable of out of the box.
Indeed, none of the methods reported do better than the majority baseline, except \method with the largest GPT-3 model.

\paragraph{Why does length normalization work?} Past work offers little explanation for why \avgll should be a successful strategy, other than the intuition that estimates are strongly length biased and require compensation. Length bias may be caused by the final softmax layer of current language models assigning too much probability mass to irrelevant options at each time-step, as noted in open-ended generation, character-level language modeling, and machine translation \cite{holtzman2020curious,al2019character, peters2019sparse}. 

Another (not mutually exclusive) argument is that length normalization may account for \textit{unconditional probability} in a similar way to \mbox{\method.} Length normalization is often measured over Byte Pair Encoding (BPE) tokens \cite{sennrich2016neural} and BPE tends to produce vocabularies where most tokens are equally frequent \cite{wang2020neural}. Recent evidence suggests that language is approximately uniformly information dense \cite{levy2018communicative, levy2007speakers, jaeger2006redundancy}. As such, length in BPE tokens may correspond  roughly to a \textit{unigram} estimate of log-probability, supposing that BPE tokens have approximately uniform unigram frequency. The adjustment made by \avgll is still somewhat different than \mbox{\method,} (division of log terms rather than subtraction) but could have a similar effect, if length and probability correlate.

\section{Discussion}
\label{sec:discussion}
Language Models are density estimation functions that assign probability to every possible string, but there are often many strings that could represent a given idea equally well. Our key observation is that a generative model assigning probability to a string that \textit{represents} a certain option isn't equivalent to selecting the \textit{concept} an option corresponds to.  We expect surface form competition anywhere that generative models are used where more than one string could represent the same concept.

\method aligns the predictions being made by the model more closely with the actual task posed by multiple choice questions: ``choose the hypothesis that explains the premise'' rather than ``generate the exact surface form of the hypothesis''. From this perspective, \method does not go far enough, because the model still cannot consider the given set of options altogether when selecting its choice. This matters when answers interact with each other, e.g., ``all of the above''.

\section{Conclusion}

We conduct a large-scale comparison of standard and recent scoring functions for zero-shot inference across all GPT-2 and GPT-3 models. We show that \method consistently outperforms previous scoring functions on a wide variety of multiple choice datasets. We also argue that compensating for \emph{surface form competition} is the cause of this boost, by demonstrating that other methods work just as well as \method when surface form competition is eliminated. 
In future work we would like to explore how surface form competition affects generation, as we hypothesize that it may be the cause of overly generic outputs under high model uncertainty.

\section*{Acknowledgments}

This work was supported in part by the ARO (AROW911NF-16-1-0121), the NSF (IIS-1562364), DARPA under the MCS program through NIWC Pacific (N66001-19-2-4031), the Natural Sciences and Engineering Research
Council of Canada (ref 401233309), and the Allen Institute for AI (AI2).
We thank Mitchell Wortsman, Gabriel Ilharco, Tim Dettmers, and 	
Rik Koncel-Kedziorski for thorough and insightful feedback on preliminary drafts.

\bibliography{custom}
\bibliographystyle{acl_natbib}
\setlength\baselineskip{12pt plus 2pt}

\clearpage


\appendix
\begin{table*}[bp]
\small
\setlength{\tabcolsep}{1.7pt}
    \caption*{Multiple Choice Accuracy on GPT-2}
    \centering
    \begin{tabular}{l|cccc|cccc|cccc|ccccc}
    Params. & \multicolumn{4}{c}{125M} & \multicolumn{4}{c}{350M} & \multicolumn{4}{c}{760M} & \multicolumn{5}{c}{1.6B} \\
     & Unc & LM & Avg &  $\textrm{PMI}_\textrm{DC}$ & Unc & LM & Avg &  $\textrm{PMI}_\textrm{DC}$ & Unc & LM & Avg &  $\textrm{PMI}_\textrm{DC}$ & Unc & LM & Avg &  $\textrm{PMI}_\textrm{DC}$ & CC \\
    \midrule
    COPA  &  0.564  &  0.610  &  \textbf{0.632}  &  0.628  &  0.558  &  0.670  &  0.660  &  \textbf{0.700}  &  0.556  &  0.698  &  0.676  &  \textbf{0.694}  &  0.560  &  0.690  &  0.684  &  \textbf{0.716}  &  -  \\
    SC  &  0.495  &  0.600  &  0.615  &  \textbf{0.670}  &  0.489  &  0.630  &  0.667  &  \textbf{0.716}  &  0.503  &  0.661  &  0.688  &  \textbf{0.734}  &  0.512  &  0.676  &  0.715  &  \textbf{0.763}  &  - \\
    HS  &  0.271  &  0.286  &  \textbf{0.295}  &  0.291  &  0.298  &  0.322  &  \textbf{0.376}  &  0.328  &  0.309  &  0.350  &  \textbf{0.432}  &  0.351  &  0.331  &  0.384  &  \textbf{0.489}  &  0.378  &  - \\
    R-M  &  0.222  &  0.361  &  0.406  &  \textbf{0.409}  &  0.213  &  0.387  &  0.420  &  \textbf{0.424}  &  0.214  &  0.393  &  \textbf{0.439}  &  \textbf{0.439}  &  0.223  &  0.415  &  0.446  &  \textbf{0.447}  &  - \\
    R-H  &  0.209  &  0.275  &  0.310  &  \textbf{0.344}  &  0.215  &  0.304  &  0.326  &  \textbf{0.363}  &  0.215  &  0.318  &  0.345  &  \textbf{0.383}  &  0.219  &  0.330  &  0.357  &  \textbf{0.391}  &  - \\
    ARC-E &  0.313  &  \textbf{0.429}  &  0.378  &  0.393  &  0.327  &  \textbf{0.494}  &  0.434  &  0.424  &  0.334  &  \textbf{0.527}  &  0.467  &  0.470  &  0.334  &  \textbf{0.562}  &  0.496  &  0.499  &  - \\
    ARC-C &  0.198  &  0.201  &  0.235  &  \textbf{0.282}  &  0.197  &  0.228  &  0.254  &  \textbf{0.286}  &  0.221  &  0.231  &  0.266  &  \textbf{0.316}  &  0.211  &  0.252  &  0.279  &  \textbf{0.338}  &  - \\
    OBQA  &  0.11  &  0.164  &  0.272  &  \textbf{0.324}  &  0.108  &  0.186  &  0.302  &  \textbf{0.386}  &  0.108  &  0.194  &  0.296  &  \textbf{0.432}  &  0.114  &  0.224  &  0.348  &  \textbf{0.460}  &  - \\
    CQA  &  0.170  &  0.255  &  0.307  &  \textbf{0.364}  &  0.165  &  0.309  &  0.352  &  \textbf{0.418}  &  0.170  &  0.333  &  0.368  &  \textbf{0.445}  &  0.171  &  0.386  &  0.385  &  \textbf{0.478}  &  - \\
    \midrule
    BQ  &  \textbf{0.622}  &  0.588  &  0.588  &  0.511  &  \textbf{0.622}  &  0.608  &  0.608  &  0.497  &  \textbf{0.622}  &  0.580  &  0.580  &  0.467  &  \textbf{0.622}  &  0.563  &  0.563  &  0.495  &  - \\
    RTE  &  \textbf{0.527} & 0.516 & 0.516 & 0.498  &  0.473  &  0.531  &  0.531  &  \textbf{0.549}  &  0.473  &  0.531  &  0.531  &  \textbf{0.542}  &  0.473  &  0.477  &  0.477  &  \textbf{0.534}  &  0.485  \\
    CB  &  0.089  &  0.482  &  0.482  &  \textbf{0.500}  &  0.089  &  \textbf{0.500}  &  \textbf{0.500}  &  \textbf{0.500}  &  0.089  &  0.482  &  0.482  &  \textbf{0.500}  &  0.089  &  \textbf{0.500}  &  \textbf{0.500}  &  \textbf{0.500}  &  0.179 \\
    SST-2  &  0.499 & 0.636 & 0.636 & \textbf{0.671}  &  0.499 & 0.802 & 0.802 & \textbf{0.862}  &  0.499 & 0.770 & 0.770 & \textbf{0.856} &  0.499 & 0.840 & 0.840 & \textbf{0.875}  &  0.820 \\
    SST-5  &  0.181 & 0.274 & 0.244 & \textbf{0.300} & 0.176 & 0.185 & 0.272 & \textbf{0.393} & 0.176 & 0.203 & \textbf{0.267} & 0.220 &  0.176 & 0.304 & 0.291 & \textbf{0.408}  &  - \\
    AGN &  0.250  &  0.574  &  0.574  &  \textbf{0.630}  &  0.250  &  0.643  &  0.643  &  \textbf{0.644}  &  0.250  &  0.607  &  0.607  &  \textbf{0.641}  &  0.250  &  0.648  &  0.648  &  \textbf{0.654}  &  0.600 \\
    TREC  &  0.226  &  0.230  &  0.144  &  \textbf{0.364}  &  0.226  &  \textbf{0.288}  &  0.122  &  0.216  &  0.226  &  0.228  &  0.226  &  \textbf{0.440}  &  0.226  &  0.228  &  0.240  &  0.328  &  \textbf{0.340} \\
    \bottomrule
    \end{tabular}
    \caption{Comparison of scoring algorithms when using GPT-2 for zero-shot inference on multiple choice questions.}
    \label{tab:mc-2}
\end{table*}

\begin{table*}[bp]
\scriptsize
\setlength\tabcolsep{2pt}
\begin{tabular}{llP{0.8\textwidth}}
    \toprule
    \textbf{Type} & \textbf{Dataset} & \textbf{Template}  \\
    \midrule
    \multirow{7}{*}{Continuation} & \multirow{2}{*}{COPA} & \premise{The man broke his toe} \dpremise{because} \uhypo{he got a hole in his sock.} \\
    \hhline{~~-} 
    & & \premise{I tipped the bottle} \dpremise{so} \uhypo{the liquid in the bottle froze.} \\
    \hhline{~--}
    & \multirow{2}{*}{StoryCloze} & \premise{Jennifer has a big exam tomorrow. She got so stressed, she pulled an all-nighter. She went into class the next day, weary as can be. Her teacher stated that the test is postponed for next week.} \dpremise{The story continues:} \uhypo{Jennifer felt bittersweet about it.} \\
    \hhline{~--}
    & \multirow{2}{*}{HellaSwag} & \premise{A female chef in white uniform shows a stack of baking pans in a large kitchen presenting them. the pans} \uhypo{contain egg yolks and baking soda.} \\
    \midrule
    \multirow{6}{*}{QA} & \multirow{3}{*}{RACE} & \premise{There is not enough oil in the world now. As time goes by, it becomes less and less, so what are we going to do when it runs out [...].} question: \premise{According to the passage, which of the following statements is true}\dpremise{?} answer: \uhypo{There is more petroleum than we can use now.} \\
    \hhline{~--}
    & ARC & \premise{What carries oxygen throughout the body?} \dpremise{the answer is:} \uhypo{red blood cells.} \\
    \hhline{~--}
    & OBQA & \premise{Which of these would let the most heat travel through?} \dpremise{the answer is:} \uhypo{a steel spoon in a cafeteria.} \\
    \hhline{~--}
    & CQA & \premise{Where can I stand on a river to see water falling without getting wet?} \dpremise{the answer is:} \uhypo{bridge.} \\
    \midrule
    \multirow{2}{*}{Boolean QA} & \multirow{2}{*}{BoolQ} & title: \premise{The Sharks have advanced to the Stanley Cup finals once, losing to the Pittsburgh Penguins in 2016 [...]} question: \premise{Have the San Jose Sharks won a Stanley Cup?} \dpremise{answer:} \uhypo{No.} \\    \midrule
    \multirow{4}{*}{Entailment} & \multirow{2}{*}{RTE} & \premise{Time Warner is the world's largest media and Internet company.} question: \premise{Time Warner is the world's largest company.} \dpremise{true or false? answer:} \uhypo{true.} \\
    \hhline{~--}
    & \multirow{2}{*}{CB} & question: Given that \premise{What fun to hear Artemis laugh. She's such a serious child.} Is \premise{I didn't know she had a sense of humor. } true, false, or neither? \dpremise{the answer is:} \uhypo{true.} \\
    \midrule
    \multirow{5}{*}{\specialcellleft{Text\\Classification}} & SST-2 & ``\premise{Illuminating if overly talky documentary}'' \dpremise{[The quote] has a tone that is} \uhypo{positive.} \\
    \hhline{~--}
    & SST-5 & ``\premise{Illuminating if overly talky documentary}'' \dpremise{[The quote] has a tone that is} \uhypo{neutral.} \\
    \hhline{~--}
    & \multirow{2}{*}{AG's News} & title: \premise{Economic growth in Japan slows down as the country experiences a drop in domestic and corporate [...]} summary: \premise{Expansion slows in Japan} \dpremise{topic:} \uhypo{Sports.} \\
    \hhline{~--}
    & TREC & \premise{Who developed the vaccination against polio?} \dpremise{The answer to this question will be} \uhypo{a person.} \\
    \bottomrule
\end{tabular}
\caption{The templates used for each task, along with an example instance (with a single random candidate answer). Original questions (premises) are colored blue, and original answers (hypotheses) are colored red. Long premises are abbreviated with ``[...]''. The full premises, conditional hypotheses and domain premises are marked in $[\cdot ]_{\text{P}}$, $[\cdot ]_{\text{UH}}$, and $[\cdot ]_{\text{DP}}$ respectively. For a complete description of our templating methodology, please see our code at \codeurl}
\label{tab:templates}
\end{table*}

\section{GPT-2 Results}
\label{sec:gpt2}

Table~\ref{tab:mc-2} shows the results for zero-shot multiple choice using GPT-2.

\section{Templates}
\label{sec:templates}

Table~\ref{tab:templates} shows an example of each template used for each dataset.

\end{document}